\documentclass[conference]{ieeeconf}
\IEEEoverridecommandlockouts    

\usepackage{multirow}
\usepackage{lipsum}
\usepackage{amsmath}
\usepackage{amssymb}
\usepackage[ruled,vlined]{algorithm2e}
\usepackage{graphicx}
\usepackage{url}
\usepackage{tabularx}
\graphicspath{{./Figures/}}
\usepackage[hidelinks]{hyperref}

\usepackage[capitalise]{cleveref}
\usepackage{tikz}
\usetikzlibrary{shapes.geometric, arrows.meta, positioning}

\title{\LARGE \bf Combining Evasive and Braking Reactions for\\  Safety Reference Models in Automated Vehicles}

\author{
	\parbox{\textwidth}{%
		\centering
		Riccardo Don\`a$^{1}$, Konstantinos Mattas$^{1}$, and Biagio Ciuffo$^{1}$%
	}%
	\thanks{$^{1}$Joint Research Centre for the European Commission, Ispra, IT
		{\tt\small riccardo.dona@ec.eurupa.eu, konstantinos.mattas@ec.europa.eu, biagio.ciuffo@ec.europa.eu}}%
}

\begin{document}
	
\maketitle
\thispagestyle{empty}
\pagestyle{empty}
	
\begin{abstract}
Computational models of careful and competent human drivers are essential for scenario-based evaluation of automated driving systems (ADS). However, most existing safety reference models primarily focus on longitudinal braking, neglecting the role of evasive steering in human collision avoidance. This paper proposes a hybrid Fuzzy-Safety Model (FSM-H) that integrates longitudinal mitigation and lateral avoidance within a unified behavioral framework. The braking component is governed by Proactive Fuzzy Safety (PFS) metrics, representing the erosion of longitudinal safety margins, while the steering component is driven by Criticality Fuzzy Safety for lane-change (CFS-LC), capturing lateral conflict severity and maneuver feasibility. A finite-state architecture models the sequential escalation from nominal driving to braking and, when necessary, to evasive steering, incorporating perception–reaction time and lane-check delays to reflect human decision processes. The model is evaluated in reconstructed high-criticality cut-in scenarios and compared with braking-only and steering-only reference strategies. Results show that the hybrid approach expands the preventability envelope while maintaining behavioral plausibility and computational tractability. The proposed framework provides a transparent and explainable human reference model suitable for simulation-based ADS safety benchmarking and regulatory assessment.
\end{abstract}

\section{Introduction}
\label{sec:introduction}
The concept of Careful and Competent (C\&C) human driver is a pivotal topic in the safety assessment of Automated Driving Systems (ADS), recognized by both the research community \cite{wang2024application} and the relevant legislation \cite{ADS}. Although it is generally acknowledged that an ADS should be at least as safe as a C\&C, the exact threshold set for the C\&C will dictate the minimum capabilities of such an ADS and its readiness determination for market introduction. 

One common usage for C\&C driver models is to run a simulation-based assessment of ADS by comparing the outcome of a simulation analysis leveraging the reference model with the results obtained from a virtual or physical experiment involving the ADS. In this regard, computational efficiency and accuracy in reproducing the relevant driving dynamics at the microscopic or mesoscopic level are key qualities a driver model should feature.

Multiple approaches have been proposed in the literature to quantitatively formalize the C\&C \cite{kitajima2025defining, wang2024application} in order to provide a suitable benchmark for an ADS. Solutions range from cognitive models \cite{arkady2022cogn,fries2022driver} to equation-based approaches \cite{mattas2022driver,jp2020FRAV}. Additionally, models are further divided into the actual reaction maneuver: either braking and/or evasive steering \cite{seiniger2013open}, and based on the type of scenarios the models can handle, \textit{e.g.}, cut-in, cut-out, or car-follow. 

\begin{figure}[!ht]
    \centering
    \includegraphics[width=\linewidth]{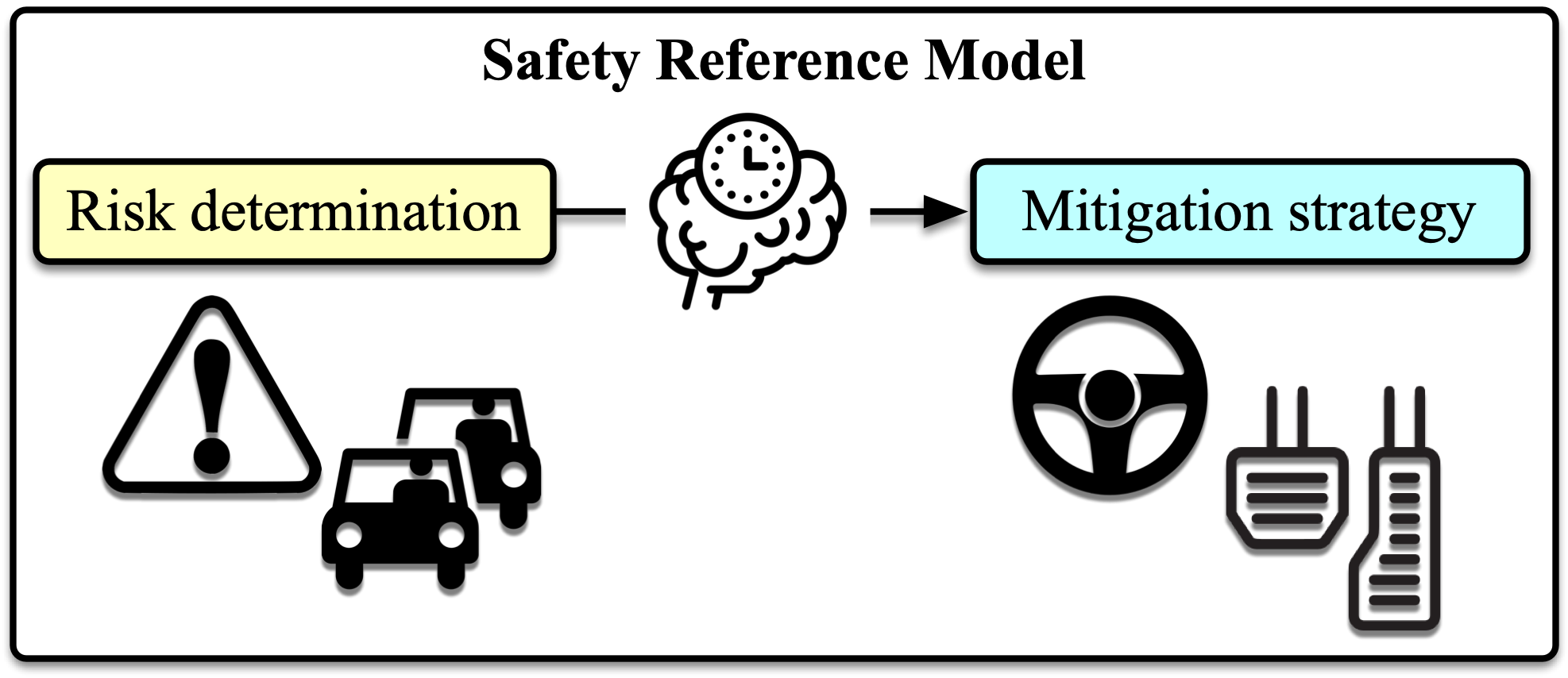}
    \caption{Safety Reference Model (SRM) general formulation}
    \label{fig:srm}
\end{figure}

Regardless of the specific realization, the general formulation for a Safety Reference Model (SRM) relies on a two-step structure shown in \cref{fig:srm}. The first step is the establishment of the instantaneous driving \textit{risk} using Surrogate Safety Metrics (SSMs) \cite{wang2021review}. Such SSMs can be existing formulations, like Time-To-Collision (TTC) or custom-defined metrics such as the Proactive/Critical fuzzy metrics in the Fuzzy Safety Model (FSM) in \cite{mattas2022driver}. The second step is the definition of a \textit{mitigation} strategy, which normally follows a certain reaction time. The most common mitigation strategy considered is the braking reaction, such as is the case for the safety reference models listed by the UN Regulation 157 \cite{157} for SAE J3016 Level 3 systems. However, evasive lane-change is also being investigated by research practitioners using both physics-based modeling approaches \cite{park2021emergency, dona2026investigating} and AI-based solutions \cite{guo2023modeling}. Eventually, state-of-the-art approaches, including Waymo's Safety Reference Model (SRM) non-impaired with eyes on the conflict (NIEON) model \cite{scanlon2022collision}, have attempted to introduce combined braking and swerving maneuvers.

Taking inspiration from a recently published manuscript introducing evasive lane-change as part of the safety envelope computation \cite{dona2026investigating} and combined solutions \cite{scanlon2022collision}, this paper aims at progressing the discussion in the C\&C human driver modeling by postulating a \emph{hybrid} braking/evasive solution. The goal is not to increase the ADS performance benchmark beyond unrealistic expectations, but to replicate the behavior observed in human drivers when approaching safety-critical situations. Crash investigation findings leveraging Event Data Recorders (EDR) reveal that human drivers usually apply some braking action before undertaking the evasive maneuver \cite{scanlon2015analysis}. 

This maneuvering strategy yields two advantages: 
\begin{enumerate}
    \item even if the collision cannot be avoided, there is at least some \emph{mitigation} through speed reduction, which would otherwise be absent in a purely evasive maneuver;
    \item reducing the speed can ease the subsequent evasive steering by making the vehicle easier to control.
\end{enumerate}

Delaying evasive steering in favor of initial braking may preserve behavioral plausibility, yet it can also degrade pure avoidance performance, revealing a non-trivial planning trade-off between early lateral displacement and progressive risk mitigation. This paper provides an initial quantitative characterization of this trade-off by extending the existing C\&C driver models at the risk determination layer while enriching the mitigation strategy with a hybrid brake–steer logic. Specifically, the Fuzzy-Safety Model (FSM) introduced by Mattas et al. \cite{mattas2022driver} and referenced in UN Regulation 157 is combined with its lane-change extension \cite{dona2023towards} to synthesize a unified braking and evasive SRM. The resulting framework preserves the modular, explainable, and computationally lightweight structure of the original FSM while enabling sequential mitigation and avoidance responses. The results show that this hybrid approach expands the modeled preventability while preserving behavioral plausibility and computational transparency. The contribution supports the development of explainable and reproducible human reference models for simulation-based ADS safety benchmarking and regulatory assessment.

\section{Design and Implementation}
\label{sec:designandimplementation}

\subsection{Fuzzy Safety Models}\label{subsec:FSM}
The model realization starts from the Fuzzy Safety Model (FSM) devised as an SRM, \emph{i.e., the performance benchmark} for the UN-R 157 for rear-end scenarios. The model was selected because of its equation-based and modular nature that allows tweaking while retaining explainability. Moreover, the model has undergone substantial validation \cite{mattas2022driver,olleja2025validation} against naturalistic datasets to further corroborate the underlying assumptions. A key feature of the FSM is its fuzzy nature that enables a smooth transition between a non-critical situation and a critical one while using a modulated reaction. This behavior is in contrast with other SRMs that normally use hard thresholds to trigger the full risk-minimizing behavior. 

The FSM model relies, in fact, on two fuzzy metrics to assess the criticality of the driving scenario based on the continuous evaluation of the relative distance and relative speed with other road users. The first one is the ``Proactive Fuzzy Safety'' (PFS), which measures the safety distance \eqref{eq:dPFSsafe} a follower should keep in order to be able to perform a full stop maneuver should the leader suddenly apply a certain braking action $b_{L, \max}$: 

\begin{equation}\label{eq:dPFSsafe}
d_{\mathrm {SAFE}}^{\mathrm {PFS}}=u_{\mathrm {ego}}\left ({{ t }}\right )\tau +\frac {u_{\mathrm {ego}}^{2}\left ({{ t }}\right )}{2b_{\mathrm {ego,comf}}}-\frac {u_{L}^{2}\left ({{ t }}\right )}{2b_{L,\max }}+d_{\mathrm {margin}}.
\end{equation}

Such a metric results in a safe car-following distance or headway policy that prevents tailgating. 

The second one is the ``Criticality Fuzzy Safety'' (CFS), 

\begin{equation}\label{eq:dCFSsafe}
d_{\mathrm{SAFE}}^{\mathrm{CFS}}
=(\Delta u)\,\tau+\frac{\left(u_{\mathrm{ego}}(t)+b_{\mathrm{ego}}(t)\,\tau-u_{L}(t)\right)^2}{2\,b_{\mathrm{ego,comf}}}
\end{equation}

which evaluates the safety distance \eqref{eq:dCFSsafe} a follower should have in order to match the instantaneous speed of a (slower) leading vehicle, assuming some acceleration constraints. The two metrics result in a safety envelope that, if violated, triggers a modulated braking reaction.  

In \eqref{eq:dPFSsafe}, \eqref{eq:dCFSsafe}, $u_{\mathrm {ego}}$ is the instantaneous longitudinal ego speed, $\tau$ the risk perception reaction time, $b_{\mathrm {ego,comf}}$ the comfortable deceleration the ego vehicle can exert, $u_{\mathrm {L}}$ is the instantaneous longitudinal leader speed, $b_{\mathrm{L,\max}}$ the assumed sudden deceleration for the leader vehicle, and $d_{\mathrm {margin}}$ the minimum margin safety distance. 

The corresponding ``unsafe'' PFS/CFS distances are obtained from \eqref{eq:dPFSsafe} and \eqref{eq:dCFSsafe} by substituting the parameter $b_{\mathrm {ego,comf}}$ with $b_{\mathrm {ego,max}}$ and removing the margin distance $d_{\mathrm {margin}}$. The parameters for the FSM models are reported in \cref{tab:model_parameters} and stem directly from UN Regulation 157 Annex 3.

\begin{figure}[!ht]
    \centering
    \includegraphics[width=\linewidth]{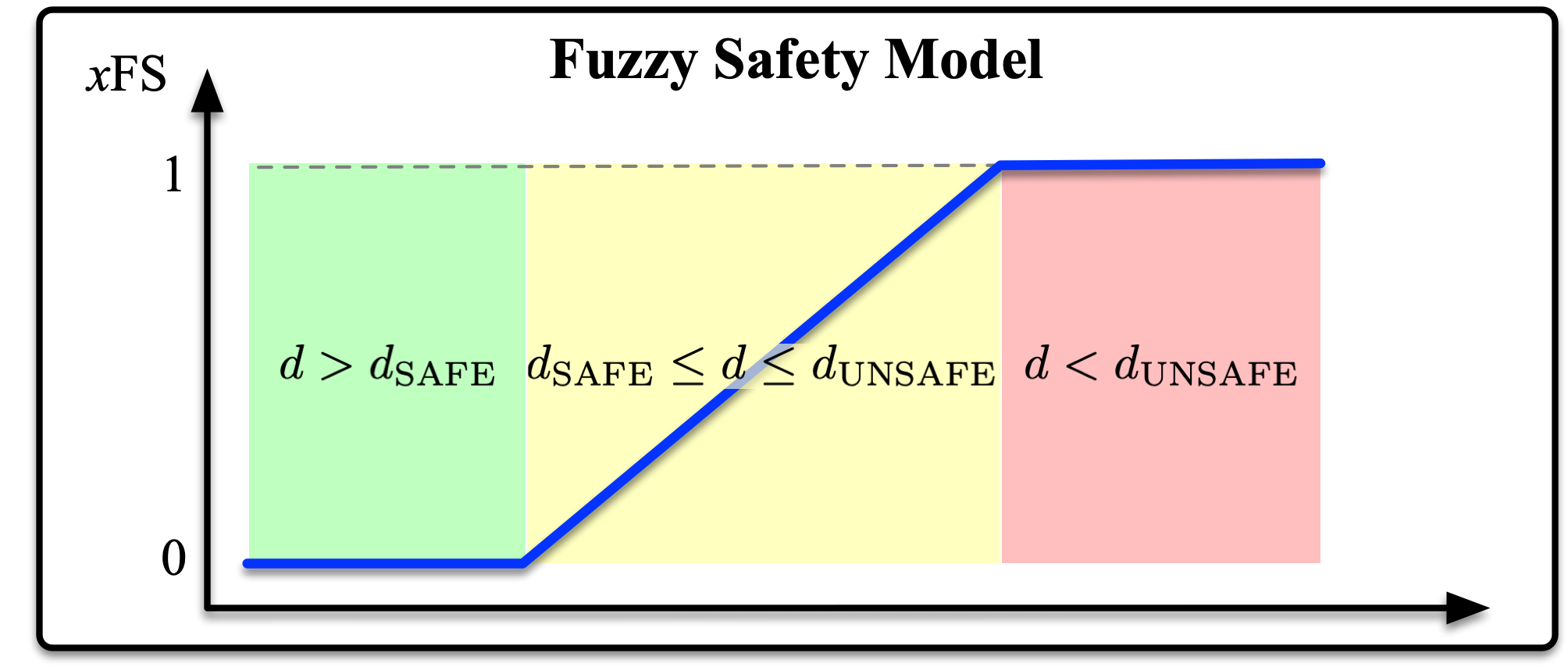}
    \caption{FSM fuzzy logic. The \textit{x}FS notation is representative of the different criticality metrics used in this paper (PFS, CFS, \dots), all of which follow the logic depicted.}
    \label{fig:fsm}
\end{figure}

The fuzzy nature is exemplified in \cref{fig:fsm}. Whenever the distance between the ego and the target vehicle is greater than the ``SAFE'' distance, the corresponding criticality metric, either the PFS or the CFS, is assigned a zero value. On the other side, when the actual distance is lower than the ``UNSAFE'' threshold, the \textit{x}FS associated with is assigned a value equal to 1. In between, the criticality metrics adopt a linear interpolation.

The amount of braking reaction modulation is dictated by the logic in \eqref{eq:rFSM}. In particular, when only the PFS is triggered, the ego vehicle can produce a deceleration effort which is in the range of 0-100\% of the $b_{ego,\text{comf}}$. Conversely, if the CFS is triggered instead, the full deceleration capability $b_{ego, \text{max}}$ can be achieved.

\begin{equation}\label{eq:rFSM}
b_{ego} =
\begin{cases}
\begin{aligned}
    &\textnormal{CFS} \cdot (b_{ego, \text{max}} - b_{ego,\text{comf}}) \\
    & + b_{ego,\text{comf}},
\end{aligned} 
& \text{if } \textnormal{CFS} \neq 0 \\[10pt]
\textnormal{PFS} \cdot b_{ego,\text{comf}}, 
& \text{if } \textnormal{CFS} = 0
\end{cases}
\end{equation}

Conversely, the ``FSM-LC'' introduced in \cite{dona2026investigating} substitutes the braking maneuver of the FSM with an evasive lane-change maneuver while replicating the same structure of the FSM concerning the criticality metrics PFS/CFS. In particular, the dual of the PFS for the FSM-LC, the ``PFS-LC'', is the safety distance it takes to perform a full lane-change maneuver in analogy to a full braking maneuver,

\begin{equation}\label{eq:cPFSLCsafe}
\begin{aligned}
d^{\mathrm{PFS}}_{\mathrm{SAFE,LC}} = \max \Big(
& u_{\mathrm{ego}}\!\left(\tau + \hat{t}_{\mathrm{LC\!-\!comf}}(t)\right) \\
& - \frac{u_L^{2}(t)}{2 b_{L,\max}}
+ d_{\mathrm{margin}},\,
0.5\,u_{\mathrm{ego}}
\Big),
\end{aligned}
\end{equation}

where $\hat{t}_{\mathrm{LC\!-\!comf}}(t)$ is the time to perform a full lane-change based on the jerk-limited kinematic model presented in \cite{soudbakhsh2011evasive}.

On the other side, the dual of the CFS for the FSM-LC,  the ``CFS-LC'', is a maneuver targeting a lateral displacement $\Delta y$ sufficient to clear the obstacle, in a similar manner to a braking maneuver that achieves the same speed as the leader vehicle

\begin{equation}\label{eq:dCFSLCsafe}
\begin{aligned}
d_{\mathrm {SAFE,LC}}^{\mathrm {CFS}} 
& =\left ({{ \Delta u }}\right )\tau  \\
&+ \left ({{ u_{ego}-u_{L}\left ({{ t }}\right ) }}\right )\hat {t}_{\mathrm {\Delta }y-\mathrm {comf}}\left ({{ y\left ({{ t }}\right ),v_{y}\left ({{ t }}\right ) }}\right ).
\end{aligned}
\end{equation}

The performance advantage of the FSM-LC lies in the much shorter safety distances with respect to the original FSM. This implies greater avoidance capabilities due to the reduced time it takes to carry out the evasive maneuver with respect to braking as the speed is increased \cite{dona2023towards}.

Eventually, following an equivalent fuzzy logic-based assignment of the criticality metrics as depicted in \cref{fig:fsm}, the target lateral acceleration is given by

\begin{equation}\label{eq:rFSMLC}
a_{y} =
\begin{cases}
\begin{aligned}
& \text{CFS}_{\mathrm{LC}}
  \big(a_{y,\mathrm{evasive}} - a_{y,\mathrm{comf}}\big) \\
& \quad + \operatorname{sign}(\text{CFS}_{\mathrm{LC}})
  \, a_{y,\mathrm{comf}},
\end{aligned}
& \text{if } \text{CFS}_{\mathrm{LC}} \neq 0, \\[10pt]
\text{PFS}_{\mathrm{LC}} \, a_{y,\mathrm{comf}},
& \text{if } \text{CFS}_{\mathrm{LC}} = 0.
\end{cases}
\end{equation}

Different from the CFS, the CFS$_\text{LC}$ can also take a negative sign to reverse the sign of the acceleration to zero the lateral speed and align the ego vehicle with the target's lane heading. 

\subsection{Hybrid Model Realization}\label{subsec:FSMH}
\begin{figure}[ht]
\vspace{10pt}
\centering
\begin{tikzpicture}[
    node distance=1.cm and 0.4cm,auto,
    startstop/.style={ellipse, draw, minimum width=2.5cm, minimum height=0.7cm, align=center, font=\small},
    decision/.style={diamond, draw, aspect=2.5, minimum width=3cm, minimum height=1cm, align=center, inner sep=0pt, font=\small},
    process/.style={rectangle, draw, rounded corners, minimum width=2.8cm, minimum height=0.8cm, align=center, font=\small},
    arrow/.style={thick, -{Stealth[scale=1.2]}}
]

\node (start) [startstop] {Compute required reaction};
\node (dec1) [decision, below=0.7cm of start] {$\tau$ \\ braking elapsed};
\node (dec2) [decision, below left=2.cm and 0.cm of dec1] {$\tau_{\mathrm{check\,lane}}$ \\ steering elapsed};
\node (proc1) [process, below right=1.9cm and 0cm of dec1] {Apply braking \\ PFS reaction};
\node (dec3) [decision, below=0.7cm of dec2] {Evasive needed};
\node (proc2) [process, right=1.cm of dec3] {Keep braking \\ PFS reaction};
\node (proc3) [process, below=1.2cm of dec3] {Apply evasive \\ CFS-LC reaction};
\node (proc4) [process, below right=1.5cm and 1.7cm of dec3] {Reset braking \\ PFS reaction};
\coordinate (split) at ([yshift=-0.6cm]dec1.south);
\coordinate (split2) at ([yshift=-0.6cm]dec3.south);

\draw [arrow] (start) -- (dec1);
\draw [arrow] (dec1.west) .. controls +(-1.,1) and +(-1.,-1) .. (dec1.west) 
    node [pos=0.5, left] {\small No};
\draw [thick] (dec1.south) -- (split) node [pos=0.4, right] {\small Yes};
\draw [arrow] (split) -| (dec2);
\draw [arrow] (split) -| (proc1);
\draw [arrow] (dec2.west) .. controls +(-1.,1) and +(-1.,-1) .. (dec2.west) 
    node [pos=0.5, left] {\small No};
\draw [arrow] (dec2) -- node [anchor=west] {\small Yes} (dec3);

\draw [arrow] (dec3) -- node [anchor=south] {\small No} (proc2);

\draw [thick] (dec3.south) -- (split2) node [pos=0.4, right] {\small Yes};
\draw [arrow] (split2) -| (proc3);
\draw [arrow] (split2) -| (proc4);

\end{tikzpicture}
\caption{FSM-H reaction mechanism.}\label{fig:FSMH}
\end{figure}
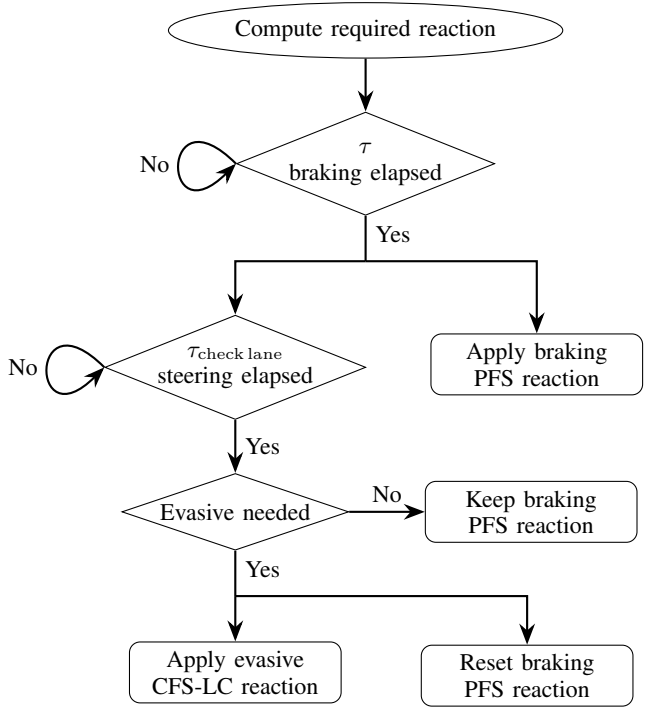

The FSM-H builds upon the existing FSM/FSM-LC structure described in \cref{subsec:FSM} by utilizing the same PFS logic of the FSM, coupled with the CFS-LC logic from the FSM-LC. The resulting mechanism is graphically shown in \cref{fig:FSMH}. 

The modeling approach is grounded in EDR findings indicating that drivers often initiate a braking action prior to committing to an evasive maneuver. Due to its proactive formulation, the PFS metric typically becomes active before the lateral feasibility metric CFS-LC, as longitudinal safety margins are generally eroded earlier than lateral maneuvering limits. This naturally results in an initial braking response. If such longitudinal mitigation is insufficient to restore a safe state, the subsequent activation of CFS-LC represents a progression toward a more critical regime, prompting the transition to evasive steering. The sequential triggering, therefore, reflects an escalation of response intensity rather than an arbitrary switching logic.

More in detail, following the determination of the needed reaction due to either one of the safety metrics exceeding zero, the FSM-H waits for a reaction time $\tau$, similarly to the other FSM variations, then it applies the braking reaction $\textnormal{PFS} \cdot b_{ego,\text{comf}}$. The triggering of the evasive steering is further awaited by an additional $\tau_{\mathrm{check\,lane}}$ representative of the time it takes for a driver to evaluate the status of the adjacent lane and originally introduced in the FSM-LC paper. 
Although a precise calibration of such a parameter was not pursued in this paper, the selected value is aligned with crash data from EDR \cite{scanlon2015analysis}.
Only after such an additional delay is any potential evasive steering undertaken based on the CFS-LC metrics. If the CFS-LC is zero, the FSM-H will retain the initially identified PFS braking policy. In case the steering action is triggered, the braking action is removed. This avoids unrealistically aggressive combined behaviors exceeding typical driver capability or dynamically unfeasible maneuvers, depending on the vehicle. 

The resulting reaction logic is thus given by the sets of equations \eqref{eq:rFSMHx} for the longitudinal dynamics and \eqref{eq:rFSMHy} for the lateral dynamics. 

\begin{equation}\label{eq:rFSMHx}
a_{x} =
\begin{cases}
0 & \text{if  }  \text{PFS} =0 \text{ and } \text{CFS}_{\mathrm{LC}} \neq 0\\[10pt]
 \textnormal{PFS} \cdot b_{ego,\text{comf}}
& \text{if }  \text{PFS} \neq 0 \text{ and } \text{CFS}_{\mathrm{LC}} = 0
\end{cases}
\end{equation}

\begin{equation}\label{eq:rFSMHy}
a_{y} =
\begin{cases}
0 & \text{if  } \text{CFS}_{\mathrm{LC}} = 0\\[10pt]
\begin{aligned}
& \text{CFS}_{\mathrm{LC}}
  \big(a_{y,\mathrm{evasive}} - a_{y,\mathrm{comf}}\big) \\
& \quad + \operatorname{sign}(\text{CFS}_{\mathrm{LC}})
  \, a_{y,\mathrm{comf}}
\end{aligned}
& \text{if } \text{CFS}_{\mathrm{LC}} \neq 0, 
\end{cases}
\end{equation}

The modular nature of the FSM/FSM-LC enables a rather straightforward realization of the FSM-H. Nonetheless, the FSM-H is still capable of producing a comparatively complex behavior as it transitions from braking to evasive steering while retaining a modulated strategy for both maneuvers. 

The parameters for the FSM-H mirror the selection adopted for the FSM and FSM-LC to ensure a fair comparison among the models and are reported in \cref{tab:model_parameters}.

\begin{table}[ht]
\centering
\caption{Model parameters for FSM, FSM-LC, and FSM-H.}
\label{tab:model_parameters}
\renewcommand{\arraystretch}{1.3}
\begin{tabular}{c||c|c|c|c}
\hline
\textbf{Param.} & \textbf{FSM} & \textbf{FSM-LC} & \textbf{FSM-H} & \textbf{Unit} \\
\hline
$\tau$                          & 0.75 & 0.75 & 0.75 & (s) \\
$\tau_{\mathrm{check\,lane}}$   & -- & 0.5 & 0.5 & (s) \\
$d_{\mathrm{margin}}$           & 2.0 & 2.0 & 2.0 & (m) \\
$a_{y,\mathrm{comf}}$           & -- & 3.0 & 3.0 & (m/s$^2$) \\
$a_{y,\mathrm{evas}}$           & -- & 5.0 & 5.0 &(m/s$^2$) \\
$b_{\mathrm{ego,comf}}$         & 4.0 & -- & 4.0 & (m/s$^2$) \\
$b_{\mathrm{ego,max}}$          & 6.0 & -- & -- &(m/s$^2$) \\
$b_{L,\mathrm{max}}$            & 7.0 & 7.0 & 7.0 & (m/s$^2$) \\
\hline
\end{tabular}
\end{table}

\subsection{Simulation benchmark suite}\label{subsec:SIM}
The high-speed cut-in scenarios from UN Regulation 157 Annex 3 were used in this paper for the analysis. The scenarios are implemented in the simulation suite openly available at \url{https://github.com/ec-jrc/JRC-FSM}.

In principle, the whole scenario selection from the UN-R 157 could be fed to the FSM-H. However, only the high-speed scenarios, \textit{i.e.}, higher than 70 km/h, are considered since at slower speeds, the evasive maneuver is no longer the most effective obstacle avoidance maneuver. A total of 75600 parameter combinations were investigated, where the ego speed ranged from 70 to 130 km/h and the cut-in vehicle speed from 10 to 120 km/h. The lateral cut-in speed for the target vehicle is contained within the 0.25 to 2.5 m/s interval, and the initial distance at which the cut-in starts lies in the 5 to 120 m range.

The sampling of the scenario parameters relies on independent uniform distributions within the mentioned intervals. This approach is intended for comprehensive parameter space exploration within the range of parameters set by UN-R 157 rather than reflecting real-world driving exposure of such cut-in maneuvers. A more realistic investigation that replicates naturalistic dynamics would require additional details concerning the operational design domain (ODD) (e.g., country of operation, road infrastructure, \dots) of the system under test. While other literature contributions, such as \cite{nakamura2022defining}, provide deeper insights into naturalistic driving distributions, characterizing such exposure is outside the scope of the current work, which focuses on the deterministic performance of the FSM-H.

\section{Results}
\subsection{Illustrative Scenario Analysis}\label{subsec:resind}
\begin{figure}[!ht]
\centering
\includegraphics[width=\linewidth]{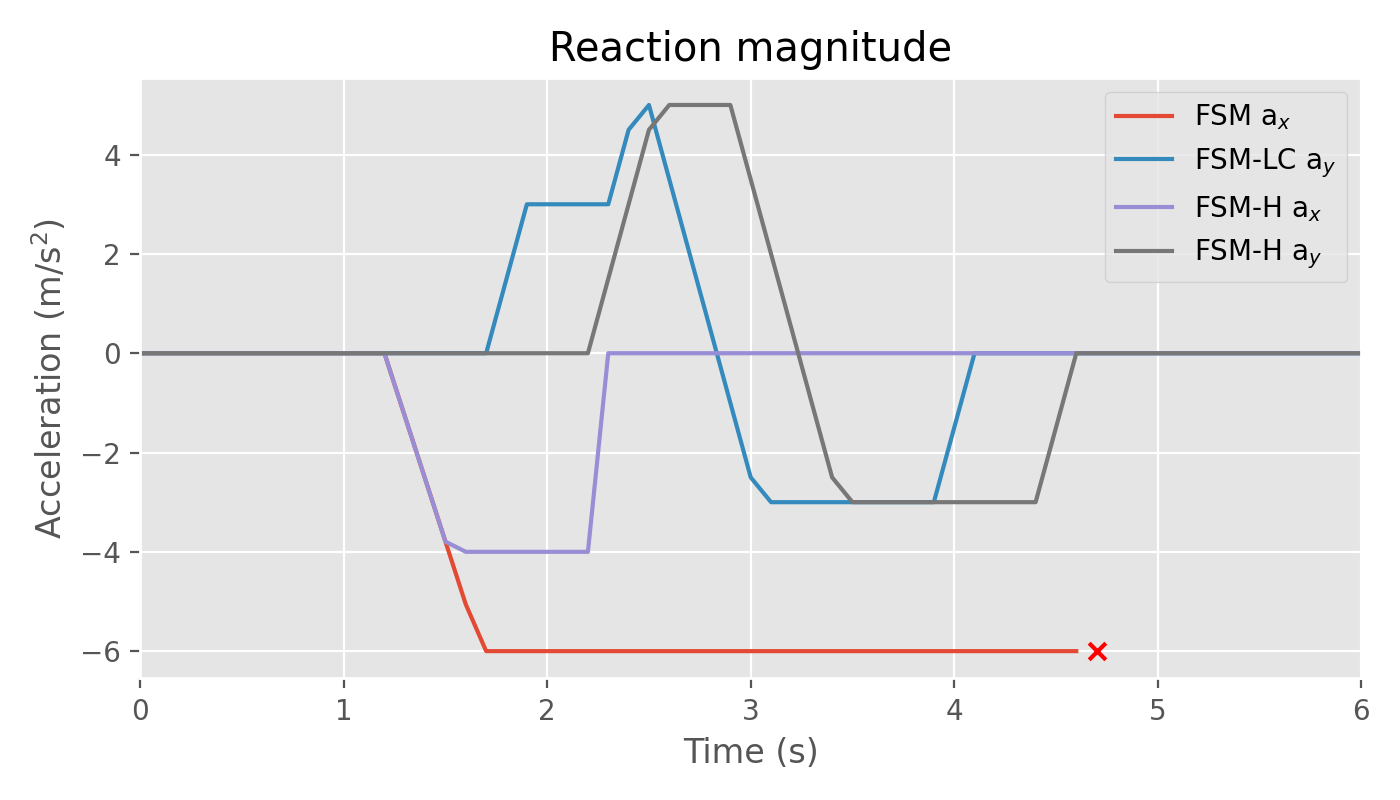}
\caption{FSM vs FSM-LC vs FSM-H cut-in analysis detailed ``hard'' cut-in scenario breakdown, criticality metrics. The scenario parameters are: $u_\text{ego} = $ 130 km/h, $u_{\text{target}} = $ 50 km/h, distance = 50 m, and lateral speed 1 m/s.  }
\label{fig:hardscn}
\end{figure}

\cref{fig:hardscn} depicts the produced accelerations by the FSM\textit{x} models investigated in an illustrative hard cut-in example scenario to elucidate the functioning of the FSM-H. The original FSM model resulted in a collision at time 4.7 s, whereas the other two models managed to safely address the scenario using, however, different strategies. The FSM-LC initially produces a mild evasive maneuver using up to 3 m/s$^2$ before the CFS-LC is eventually triggered, thus increasing the lateral acceleration effort up to 5 m/s$^2$. Conversely, the FSM-H initially starts with a $-4$ m/s$^2$ deceleration in line with FSM. Nonetheless, at 2.2 s, the CFS-LC is triggered, resetting the braking action in favour of initiating the evasive maneuver. 

\subsection{Aggregated Performance Comparison}\label{subsec:resaggr}
\cref{tab:crash_stats} reports the results of the cut-in scenarios simulation analysis for the FSM, FSM-LC, and the novel FSM-H as a function of the initial speed. The number of cases increases with the ego vehicle speed since, at higher speeds, there are more valid combinations to investigate. Indeed, one of the conditions for the cut-in scenario to be critical is for the target speed to be lower than the ego-speed.  

\begin{table}[ht]
\centering
\caption{Crash statistics by initial ego speed.}
\renewcommand{\arraystretch}{1.3}
\begin{tabular}{c||c|c|c|c}
\hline
\textbf{Speed (km/h)} & \textbf{Cases} & \textbf{FSM} & \textbf{FSM-LC} & \textbf{FSM-H}  \\
\hline
70  & 7200  & 347  & 213  & 258 \\
80  & 8400  & 506  & 287  & 344 \\
90  & 9600  & 713  & 367  & 436 \\
100 & 10800 & 979  & 456  & 538 \\
110 & 12000 & 1314 & 553  & 650 \\
120 & 13200 & 1722 & 656  & 771 \\
130 & 14400 & 2217 & 884  & 899 \\
\hline\hline
Cumulative & 75600 & 7798 & 3416 & 3896  \\
\hline
\end{tabular}
\label{tab:crash_stats}
\end{table}

The performance of the FSM-H is slightly lower than that of the purely evasive FSM-LC. Nonetheless, FSM-H delivers a higher benchmark than the regular FSM. The overall crash reduction rate is down 50\% compared to the FSM, although $\approx$ 14\% higher than the FSM-LC. 

In 43831 cases (58\% of the total), the FSM-H managed to achieve a safe state with only the PFS, thus not requiring any evasive steering maneuver. This is an important consideration as lane-change might not always be feasible or desirable. In 13766 cases (18\%), the FSM-H required the evasive steering, which is substantially lower than the FSM-LC that produced 20628 evasive maneuvers in the considered scenarios. 

Indeed, for some parameter combinations, no reaction is needed. That is the case for cut-in taking place at low lateral speed and at a closer distance, where the ego vehicle will overtake the target vehicle without necessarily taking any reaction. 

\begin{figure}[!ht]
\centering
\includegraphics[width=\linewidth]{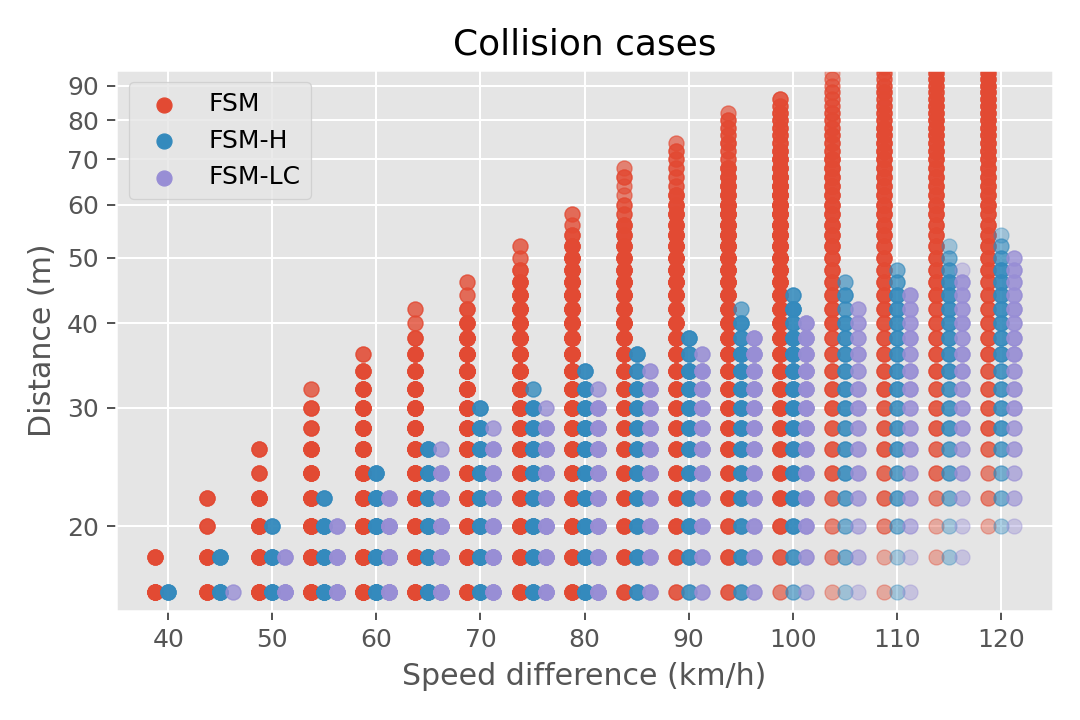}
\caption{Collision kernels density comparison.}
\label{fig:kde}
\end{figure}

\cref{fig:kde} graphically reports the collision kernel densities for the FSM, FSM-LC, and FSM-H as a function of the initial distance at which the cut-in takes place and the relative speed between the vehicles. Overall, the FSM-LC and FSM-H show a very similar behavior in terms of effectiveness of reducing the chances of high relative speed collisions versus the original FSM. The additional crashes returned by the FSM-H can be ascribed to cases where the initial criticality was so high that the original delay of the evasive steering maneuver proved detrimental to the overall safety.

\begin{figure}[!ht]
\vspace{4pt}
\centering
\includegraphics[width=\linewidth]{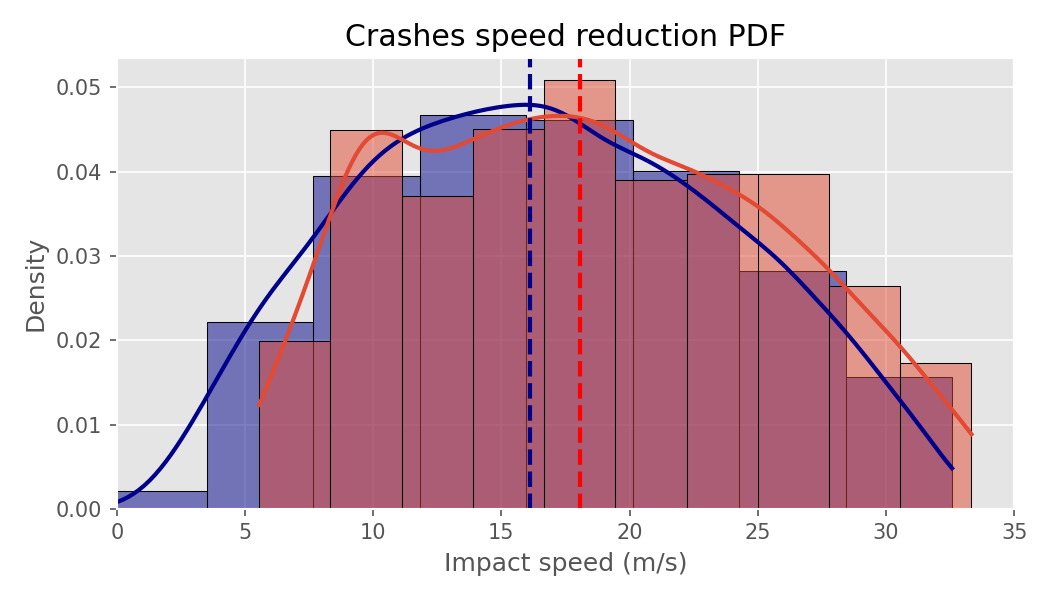}
\caption{Distribution of impact speeds for the collision scenarios.}
\label{fig:mitigation}
\end{figure}

However, \cref{fig:kde} does not convey the crash mitigation capabilities of the different models, which are instead depicted in \cref{fig:mitigation} for the FSM-H and FSM-LC.
\cref{fig:mitigation} displays the probability density function of the impact speed for the FSM-H (blue curve) and FSM-LC (orange curve). The curves are represented here in terms of density rather than frequency to ease visualization, since the FSM-LC has fewer crashes than the FSM-H. 

Indeed, as the FSM-H can perform braking, it can still try to mitigate the severity of the collision when the same is unavoidable. This mitigation capability results in a median impact speed of 16.1 m/s versus the median impact of 18.1 m/s for the FSM-LC. The result is not unexpected as it aligns with the additional reaction time $\tau_{\text{check\, lane}} =$ 0.5 s and the maximum braking deceleration of $b_{\mathrm{ego,comf}}$ of 4 m/s$^2$.

\subsection{Switching Dynamics and Escalation Behavior}\label{subsec:resswitch}
\begin{figure}[!ht]
\centering
\includegraphics[width=\linewidth]{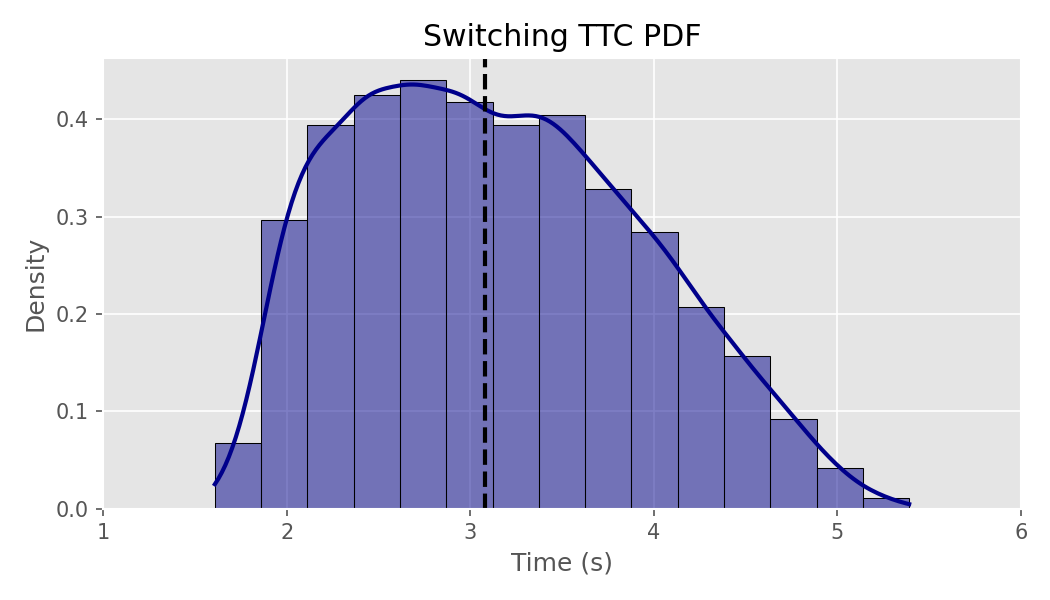}
\caption{Distribution of TTC when FSM-H switches from braking to steering. In black, the median value.}
\label{fig:TTCswitch}
\end{figure}

\cref{fig:TTCswitch} displays the probability density function of the TTC when the FSM model decides to switch from braking to evasive steering. The median value is found at 3.08 s (mean $=$ 3.14 s, $Q_1 =$ 2.50 s, and $Q_3 =$ 3.72 s), which closely aligns with the findings of \cite{kitajima2025defining}. In the latter paper, the authors found that at a TTC $<$ 3.0 s, most human drivers decided to switch from braking to evasive steering in line with the strategy of FSM-H. Notably, no specific tuning was carried out to obtain this behavior. 

\section{Summary and Conclusions}\label{sec:conclusion}
This paper presented a hybrid braking–steering Safety Reference Model (FSM-H) extending the Fuzzy Safety Model used in the context of UN Regulation 157. By combining the longitudinal Proactive Fuzzy Safety (PFS) logic with the lateral Criticality Fuzzy Safety (CFS) lane-change formulation, the proposed model enables a sequential mitigation strategy that reflects observed human driving behavior in safety-critical situations.

The results highlight the inherent trade-off between immediate evasive steering and sequential braking–steering strategies. While the purely evasive FSM-LC achieves the lowest overall crash rate, the proposed FSM-H delivers comparable avoidance performance while introducing an important mitigation capability, and, moreover, a more coherent behavior with EDR findings. The slight increase in crash frequency observed in FSM-H is primarily attributable to the additional delay $\tau_{\text{check\, lane}}$, which reduces the time window available for lateral displacement in highly critical situations. However, this reduction in preventability is partly compensated by the speed reduction achieved during the initial braking phase.

The impact-speed analysis confirms this mitigation effect. Even in scenarios where collision avoidance is not feasible, FSM-H systematically reduces kinetic energy prior to impact, resulting in a lower median impact speed compared to FSM-LC. From a behavioral perspective, the FSM-H therefore better captures the sequential decision-making process of careful and competent drivers, who typically attempt partial mitigation before committing to a full lateral maneuver.

The distribution of the Time-To-Collision (TTC) at which FSM-H transitions from braking to steering provides additional insight into this sequencing behavior. The median switching TTC of approximately 3.0 s aligns with previously reported thresholds for driver intervention in critical scenarios. This suggests that the hybrid logic does not introduce arbitrary switching behavior, but instead reproduces timing consistent with observed human responses.

Overall, the hybrid formulation reveals that the boundary between preventable and unpreventable collisions is not solely determined by the feasibility of a single optimal maneuver. Instead, it emerges from the interaction between reaction time, mitigation sequencing, and dynamic constraints. The FSM-H therefore offers a more behaviorally plausible characterization of the safety envelope compared to single-maneuver reference models.

Simulation results on high-speed cut-in scenarios indicate that FSM-H substantially improves performance compared to braking-only reference models and achieves avoidance capabilities close to those of purely evasive strategies. Importantly, the hybrid approach preserves impact-speed mitigation in cases where avoidance is not achievable, thereby offering a more comprehensive safety characterization that accounts for both preventability and severity reduction.

The proposed formulation retains the modular, explainable, and low computational structure of the original FSM, making it suitable for large-scale scenario-based ADS assessment and regulatory use. At the same time, the results underline the importance of explicitly modeling maneuver sequencing when defining the benchmark of a careful and competent human driver.

Future work will focus on extending the analysis to additional conflict typologies, investigating parameter sensitivity (e.g., reaction times and lane-check delays), and validating the hybrid switching logic against naturalistic driving data. Such efforts are necessary to further consolidate the role of hybrid safety reference models in regulatory and performance benchmarking contexts.

\section*{ACKNOWLEDGMENTS}
The work was supported by the Joint Research Centre for the European Commission. The opinions expressed are those of the authors and should not be considered to represent an official opinion of the European Commission.

\bibliographystyle{IEEEtran}
\bibliography{root} 
	
\end{document}